\documentclass[10pt,twocolumn,letterpaper]{article}

\usepackage{iccv}
\usepackage{times}
\usepackage{epsfig}
\usepackage{graphicx}
\usepackage{amsmath}
\usepackage{amssymb}
\usepackage{colortbl}
\usepackage{xcolor}
\usepackage{bm}
\definecolor{Gray}{gray}{0.9}
\usepackage{color}

\usepackage{algorithmic}
\usepackage{algorithm}
\usepackage{multirow}

\newcommand{\bx}{{\bf x}}

\newcommand{\bp}{{\bf p}}
\newcommand{\br}{{\bf r}}

\renewcommand{\paragraph}[1]{\noindent\textbf{#1}\quad}
\usepackage[accsupp]{axessibility}  % Improves PDF readability for those with disabilities.
\usepackage[breaklinks=true,bookmarks=false]{hyperref}

\iccvfinalcopy % *** Uncomment this line for the final submission

\ificcvfinal\fi

\begin{document}

%%%%%%%%% TITLE
\title{MixBag: Bag-Level Data Augmentation for Learning from Label Proportions}

\author{
Takanori Asanomi$^1$, Shinnosuke Matsuo$^1$, Daiki Suehiro$^{1,2}$, Ryoma Bise$^1$ \\
$^1$Kyushu University, Fukuoka, Japan \quad \quad $^2$RIKEN AIP, Japan\\
{\tt\small \{takanori.asanomi@human., shinnosuke.matsuo@human., suehiro@, bise@\}ait.kyushu-u.ac.jp}}
\maketitle
%%%%%%%%% ABSTRACT
\begin{abstract}
Learning from label proportions (LLP) is a promising weakly supervised learning problem. In LLP, a set of instances (bag) has label proportions, but no instance-level labels are given.
LLP aims to train an instance-level classifier by using the label proportions of the bag.
In this paper, we propose a bag-level data augmentation method for LLP called MixBag, based on the key observation from our preliminary experiments; that the instance-level classification accuracy improves as the number of labeled bags increases even though the total number of instances is fixed. 
We also propose a confidence interval loss designed based on statistical theory to use the augmented bags effectively.
To the best of our knowledge, this is the first attempt to propose bag-level data augmentation for LLP.
The advantage of MixBag is that it can be applied to instance-level data augmentation techniques and any LLP method that uses the proportion loss.
Experimental results demonstrate this advantage and the effectiveness of our method.
\end{abstract}

%%%%%%%%% BODY TEXT
\section{Introduction}
% LLPの問題設定の説明
In general classification tasks, a model is trained on datasets where each piece of data (instance) has class labels.
However, instance-level annotations cost a lot, and there are many cases in which instance-level annotation cannot be disclosed to the public for privacy reasons~\cite{qi2016learning}. In such a situation, we only know the label proportions given to a set of instances (bag) and must train a model from these label proportions to classify each instance. This problem setting is known as learning from label proportions (LLP)~\cite{ardehaly2017co}.

\begin{figure}[t]
    \centering
    \includegraphics[width=1.0\linewidth]{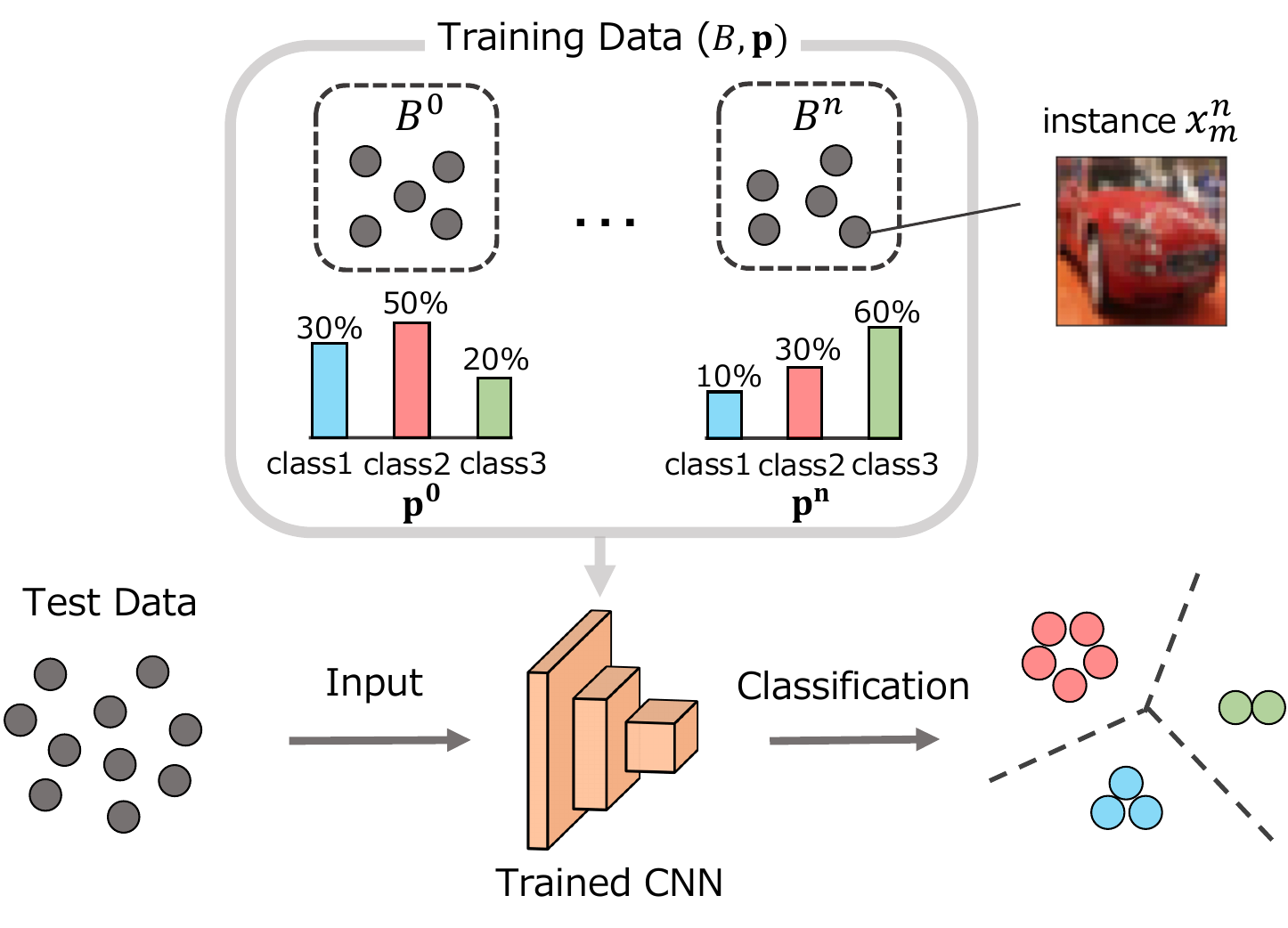}
    \caption{Illustration of learning from label proportions. 
    LLP aims to train an instance-level classifier by using label proportions of a bag as training data instead of instance-level labels. 
    }
    \label{fig:LLP}
\end{figure}

% LLPの問題設定の説明
Figure \ref{fig:LLP} illustrates the problem setup of LLP.
A bag $B$ consists of instances (e.g., images).
Bag-level labels are given as training data instead of instance-level labels (class labels of each instance $\bx$ are unknown).
The label proportion $\bp^i$ of bag $B^i$ is given as a bag-level label. For example, if a bag $B^0$ contains 60, 100, and 40 instances of class 1, 2, and 3, respectively, $\bp^i=(0.3, 0.5, 0.2)^{\mathrm{T}}$.
LLP aims to train an instance-level classifier by using the label proportions of a bag. It is a weakly supervised learning task.

Many LLP methods have been proposed to address this challenging task~\cite{ardehaly2017co,rueping2010svm,tsai2020,tokunagaECCV2020,ijcai2021p377,matsuoICASSP2023}. Most of them are based on the {\em proportion loss}, which evaluates the difference between the given proportion $\bp^i$ and the estimated proportion $\hat{\bp}^i$, and can be calculated by taking the average of the class probabilities of the instances in a bag.
The methods based on the proportion loss work well when bag-level labels (label proportions) are sufficient.
However, the accuracy decreases as the number of labeled bags decreases, i.e., insufficient labeled data causes difficulty in training a network, which is known in machine learning tasks.

In such cases, instance-level data augmentation is often an effective way to improve accuracy.
Many instance-level data augmentation techniques, such as perturbation, for general classification tasks have been proposed, and their effectiveness has been demonstrated~\cite{shorten2019survey}. 
Instance-level data augmentation may improve the accuracy in LLP.
However, it does not increase the number of labeled bags from the original one, and it is difficult to generate various bags with different proportions without using instance-level labels (these are unknown in LLP).

This paper proposes a {\em bag-level} data augmentation for LLP that can generate new bags with various label proportions.
To design an effective {\em bag-level} data augmentation technique, we conducted a detailed analysis of what situation can improve the accuracy in LLP. As a result, we found an important observation that the accuracy improves as the number of labeled bags increases, even though the total number of instances is fixed, i.e., different bags overlap each other as shown in Figure \ref{fig:overlap} (Right).
On the basis of this observation, we propose a {\em bag-level} data augmentation called MixBag.
MixBag increases the number of labeled bags artificially by sampling instances from a pair of original bags and mixing them: this operation mimics the above observations.
The expected label proportion of a generated mixed bag can be calculated by those of the original bags. However, it may have a gap from the actual proportion of the mixed bag because randomly selected instances do not follow the proportion of the original ones. This gap adversely affects the training.

We thus introduce a confidence interval loss that helps to train a classification network by using the generated bags while avoiding this adverse effect by a proportion gap; it is statistically guaranteed.
To the best of our knowledge, this is the first attempt to propose {\em bag-level} data augmentation for LLP. 
Experiments using eight datasets demonstrate the effectiveness of our method in various cases; our method improved the classification accuracy on all datasets.
Our main contributions are summarized as follows.
\begin{itemize}
    \item We examined how the number of labeled bags and the bag size affect the performance in LLP. From the preliminary experiments, we concluded that the accuracy improves as the number of labeled bags increases, even though the total number of instances is fixed.
    \item We proposed MixBag with a confidence interval loss. This method artificially increases the number of labeled bags, and the confidence interval loss helps train a classification network using the generated bags. 
    \item We demonstrated that MixBag can be applied to any of the current LLP methods and any of the instance-level augmentation methods. Experimental results show that our method improved classification accuracy on various datasets.
\end{itemize}

%-------------------------------------------------------------------------
\section{Related work}
\subsection{Learning from Label Proportion (LLP)}
Proportion loss~\cite{ardehaly2017co} is the most common approach to LLP.
To compute the proportion loss, the average of the output probability in a bag is calculated as a proportion estimation, and then the bag-level cross-entropy between the true label proportion and the estimated proportion is used as the loss.

The proportion loss has been extended in many methods; e.g., introducing regularization terms~\cite{ShiY2020,ardehaly2017co,DulacArnoldG2020,tsai2020,yang2021two,liu2019learning, ijcai2021p377}.
Tsai {\it et al.} ~\cite{tsai2020} incorporated a consistency regularization into the proportion loss, which forces the network predictions to be consistent when its input is perturbed.
Yang {\it et al.} ~\cite{yang2021two} use contrastive learning for pre-training in a self-supervised learning manner and then train the model by using the proportion loss.
LLP-GAN ~\cite{liu2019learning} introduces the generative adversarial network (GAN) to LLP. The fake instances created by the generator encourage the discriminator to distinguish real or fake and classify the real instances. This method also uses the proportion loss to train the classifier.
Two-stage LLP ~\cite{ijcai2021p377} introduces pseudo-labeling after pre-training by using a proportion loss.

It is known that the classification accuracy decreases when the labeled data is insufficient in machine learning.
To deal with this issue, data augmentation is a standard approach.
However, {\em bag-level} data augmentation for LLP is still an unexplored area.
This paper proposes a simple but effective {\em bag-level} data augmentation method. One of the advantages of our method is that any LLP method can be applied after our {\em bag-level} augmentation.

\begin{figure}[t]
    \centering
    \includegraphics[width=1.0\linewidth]{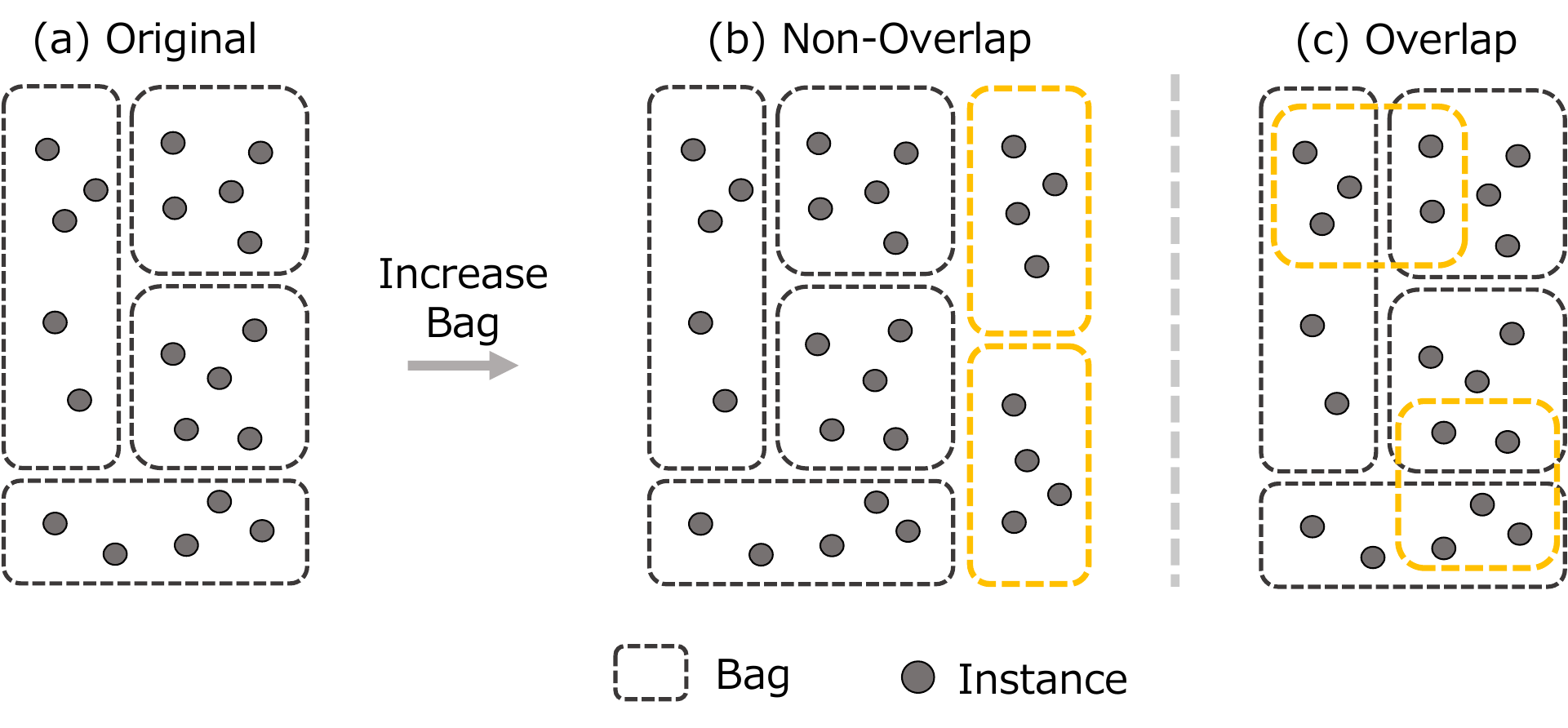}
    \caption{Illustration of overlap and non-overlap situations. (a) Original bags. (b) New bags (yellow) are added with non-overlap, i.e., the total number of instances increases. (c) New bags overlap with the original ones, i.e., the total number of instances is fixed.
    }
    \label{fig:overlap}
    \vspace{-0.5mm}
\end{figure}

\subsection{Instance-level data Augmentation}
Data augmentation is a standard way to address the problem of inadequately labeled data. Many instance-level data augmentation techniques have been proposed.
Basic data augmentation involves adding perturbations to the original images, such as by flipping the image~\cite{shorten2019survey}, adding noise~\cite{shorten2019survey}, and erasing parts of an image ~\cite{devries2017improved, chen2020gridmask, li2020fencemask}.
Image mix~\cite{inoue2018data, zhang2017mixup, yun2019cutmix, hendrycks2019augmix} is another approach, where methods such as MixUP~\cite{zhang2017mixup}and CutMix~\cite{yun2019cutmix} mix together two original images.
GAN-based augmentation methods~\cite{perez2017effectiveness, zhu2017unpaired, tang2020unified} represent data distributions and generate new images based on these distributions.
Auto augment~\cite{cubuk2019autoaugment, lim2019fast, cubuk2020randaugment} automatically searches for suitable augmentation approaches that improved performance.
The above methods are instance-level data augmentations.
It is not a simple task to generate various bags that have different proportions without using instance-level labels.
For example, CutMix and MixUp require the labels of the original images (instances) in order to prepare the label proportions of a generated new bag.
Since simple perturbation methods do not require the ground truth of the image label for augmentation, they can be used to add perturbations to each instance in a bag.
However, these methods are not designed for {\em bag-level} data augmentation.

\subsection{Bag-Level data augmentation for MIL}
Few {\em bag-level} data augmentation methods have been proposed for multiple instance learning (MIL), which trains a classifier from bag-level class labels (positive or negative).
Li et al.~\cite{li2021novel} proposed a MIL-based method that uses {\em bag-level} augmentation for COVID-19 severity assessment.
This method takes a pseudo-labeling approach, where positive and negative instances are taken from positive bags based on the estimated confidence of the instance class and used to make a new bag.
Yang et al.~\cite{yang2022remix} proposed a MIL method for segmenting pathology images. This method generates new bags by clustering instances.
It is reasonable for MIL to make a positive or negative bag from high confidence positive or negative instances.
However, it is difficult to make a proportion label from pseudo labeling since all of the pseudo labels have to be accurate to calculate the correct proportion.
So in order to improve the performance of LLP, appropriately designing a {\em bag-level} data augmentation is required.

\section{Preliminary: LLP and Proportion Loss}
In LLP, the training set contains $n$ bags, $B^1, \ldots, B^n$. Each bag $B^i$ consists of a set of instances, i.e., $B^i=\{x_j^i\}_{j=1}^{|B^i|}$, and a vector of label proportions $\bp^i=(p^i_1,\ldots,p^i_c,\ldots,p^i_C)^T$ is attached to each bag as training label, where $C$ is the number of classes.
This label indicates that $p^i_c \times |B^i|$ instances in the $i$-th bag belong to the class $c$ for any $c \in \{1,\ldots,C\}$.
The goal of LLP is to train a network that estimates the class label of each instance by using only the proportion labels of the bags.
Proportion loss~\cite{ardehaly2017co} is a standard approach to this challenging task, and many methods are based on it.

Given training data with label proportions, the standard proportion loss is a bag-level cross-entropy between the ground-truth of the label proportion $\bp^i$ and the predicted proportion $\hat{\bp}^i$, which is the average of the output class probabilities in a bag. 
It is defined as:
\begin{align}
    \ell_{\mathrm{prop}}(B^i, \bp^i, f) = - \sum_{c=1}^{C} p_c^i \log \hat{p}_c^i, \\
    \hat{p}_c^i = \frac{1}{|B^i|} \sum_{j=1}^{|B^i|} f(x_j^i)_c,
\end{align}
where $f(x_j^i)_c$ is the output probability that $x_j^i$ belongs to $c$.

\section{Preliminary Empirical Analysis of LLP}
In previous works, the relationship between the number of labeled bags, bag sizes, and accuracy in LLP has not been analyzed well. Some papers~\cite{liu2019learning, DulacArnoldG2020} reported that the accuracy increased as the larger bag size in their experiments. However, the reason was not discussed well. In their experiments, they changed the bag size while the total number of instances was fixed. Since an instance generally does not belong to two bags (i.e., bags does not overlap each other) in LLP, it indicates that the bag size and the number of labeled bags have a trade-off; when the bag size increases, the number of labeled bags decreases.
%We guess that the accuracy decreased because the number of labeled bags (labels) decreased but not by the increasing bag sizes.
%It is reasonable because the number of labeled data is generally related to the accuracy in any machine learning tasks.

In this section, we further empirically analyze the relationships among the number of labeled bags, bag sizes, and accuracy by using the proportion loss on various public datasets. 
First, we conducted experiments on whether the number of labeled bags or the bag size has more influence on the classification accuracy in a general setup of LLP, where different bags do not contain the same instance (no overlap), as shown in Figure \ref{fig:overlap} (b).
To avoid the trade-off between the number of labeled bags and the bag size, we fixed the number of labeled bags to analyze the effect for accuracy by the number of labeled bags, and vice versa. In the experiments, the total number of instances changed.
We speculated that this analysis may give clues for developing a new method in LLP.

\vspace{5pt}
\noindent
{\bf Datasets and experimental settings}: 
We used eight datasets in our experiments\footnote{Almost all papers on LLP use CIFAR10 and SVHN}: 1) CIFAR10~\cite{cifar10}, which is commonly used in classification tasks. The dataset contains ten classes, such as vehicles and animals; 2) SVHN~\cite{svhn}, which is a house-number-images dataset, and a character-level label is given like in MNIST~\cite{mnist}; 3) OCT, which contains four classes of Retinal OCT images; 4) BLOOD, which contains eight classes of Blood Cell Microscope images; 
%\red{5) ORGANA, which contains 11 classes of abdominal CT images; 6) ORGANC, which contains eleven classes of abdominal CT images; 7) ORGANS, which contains eleven classes of abdominal CT images;} 
5) ORGANA, 6) ORGANC, 7) ORGANS, which contain eleven classes of abdominal CT images. The difference is the view of the image and its sizes,
and 8) PATH, which contains nine classes of colon pathology images. Note that 3) to 8) are datasets in MedMNIST~\cite{medmnistv2}, and we did not use the small datasets in ~\cite{medmnistv2} since they cannot be used to prepare enough bags for LLP. 
%3\textasciitilde8) MedMNIST~\cite{medmnistv2}, which include 12 datasets for 2D. All images resolution are preprocessed 28$\times$28 and given correct classification labels. This dataset has a variety of biomedical images and is independent of each 12 datasets. We choose six datasets from MedMNIST because small datasets are not suitable for our task.

\begin{figure}[t]
    \centering
    \includegraphics[width=1.0\linewidth]{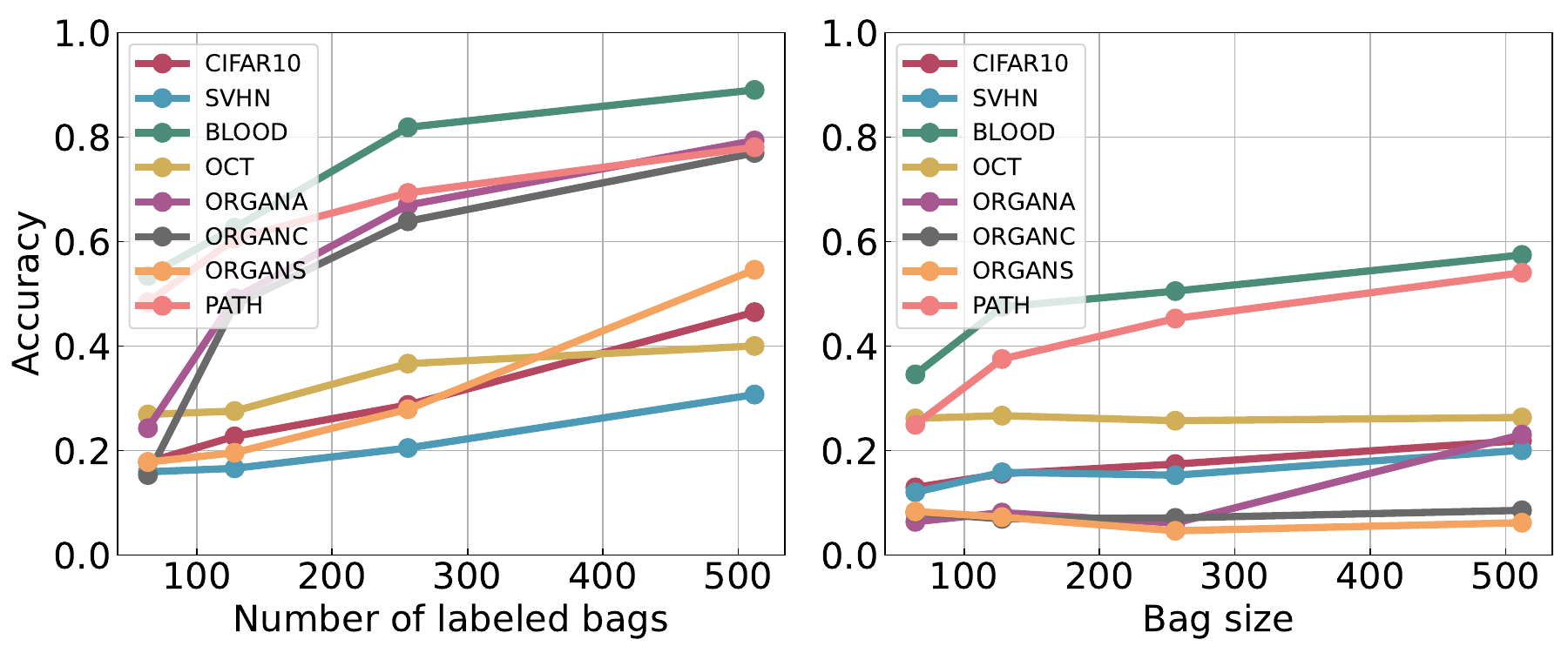}
    \caption{Relationship among the number of labeled bags, bag sizes, and accuracy in a general setup of LLP, where different bags do not contain the same instance (no overlap). {\bf Left}: Accuracy improvement when increasing the number of labeled bags (bag size is fixed). {\bf Right}: Accuracy improvement when increasing the bag size (the number of labeled bags is fixed).
    %Accuracy \red{when the overall sample is variable.} {\bf Left}: Accuracy improvement when increasing the number of labeled bags (bag size is fixed). {\bf Right}: Accuracy improvement when increasing the bag size (the number of labeled bags is fixed).
    }
    \label{fig:preliminary1}
\end{figure}

To prepare a bag, the proportions of each class were randomly selected from a Gaussian distribution; the decided proportion was used as the label of the bag. Then, samples were randomly selected from each class, in which the number of samples for each class followed the proportion of the class.
In real applications, different bags do not contain the same instance; i.e., they do not overlap. We thus generated bags without overlap in these experiments.

\vspace{5pt}

\noindent
%{\bf Performance when changing the number of labeled bags and the bag size in the case of no overlap}: 
{\bf Performance when changing the number of labeled bags in the general setup of LLP (no overlap)}:
% 〇全体サンプルが可変の場合
%・バッグ数固定、バッグサイズを変えた際の精度変化
%・バッグサイズ固定、バッグ数を変えた際の精度変化
We examined how the number of labeled bags affects the instance-level classification accuracy for the standard proportion loss. We varied the number of labeled bags (64, 128, 256, 512) while keeping the bag size fixed at 10.
In this experiment, we follow the general setup of LLP; different bags do not overlap, i.e., the number of instances changed when the number of labeled bags was changed.
As shown in Figure\ref{fig:preliminary1} (Left), the accuracy improved due to increasing the number of labeled bags in all datasets. This is reasonable because the number of labeled bags indicates the number of proportion labels; more labeled data for training is expected to improve accuracy in machine learning. 
\vspace{5pt}

\noindent
{\bf Performance when changing the bag size in the general setup of LLP (no overlap)}:
We analyzed how the bag size affects the accuracy in the general setup of LLP. In this experiment, we changed the bag size (64, 128, 256, 512) but fixed the number of labeled bags at 10.
As shown in Figure \ref{fig:preliminary1} (Right), the accuracy did not vary significantly with the bag size in six of the datasets, while it improved in two datasets.
This result shows that increasing the number of instances without increasing the labels does not effectively improve accuracy.
\vspace{5pt}

\begin{figure}[t]
    \centering
    \includegraphics[width=0.58\linewidth]{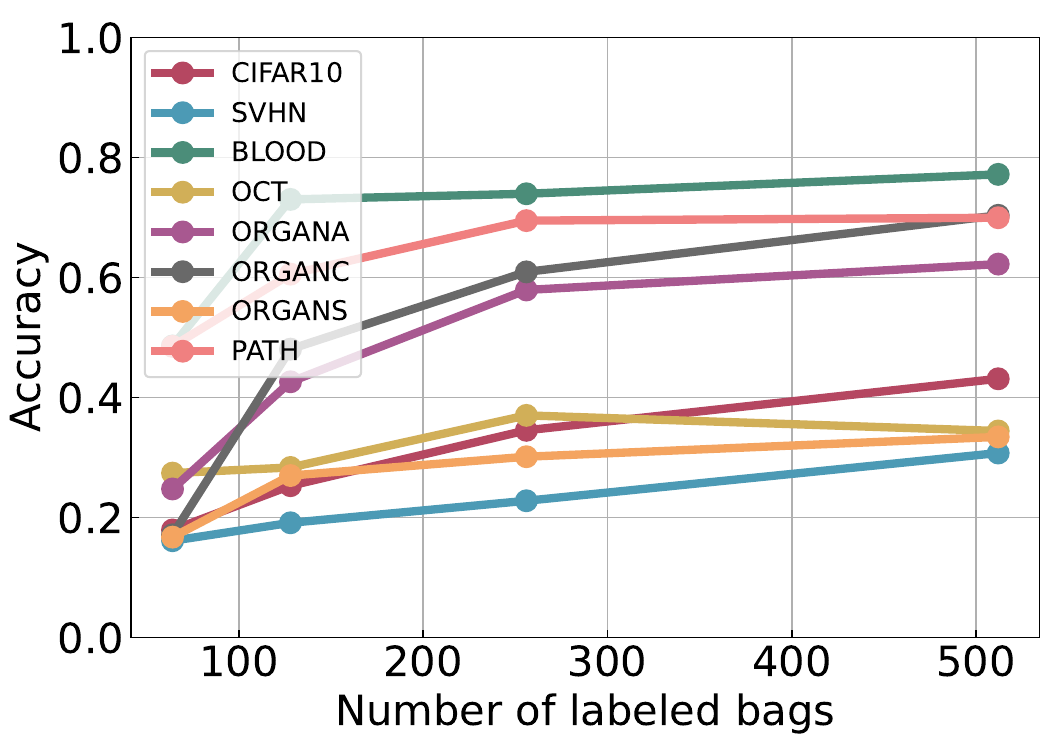}
    \caption{Experimental results on how the number of labeled bags affects the accuracy in the case of overlap, i.e., when the total number of instances was fixed.
    %Accuracy \red{when overall sample is limited.} Accuracy improvement when increasing the number of labeled bags (bag size is fixed).
    }
    \label{fig:preliminary2}
\end{figure}

\noindent
{\bf Performance when changing the number of labeled bags in the case of overlap}:
The above observations indicate that increasing the number of labeled bags is important to improving accuracy. 
Another factor that might have improved the accuracy is increasing the total number of `instances,' which occurs when the number of labeled bags increases.
To confirm whether the number of labeled bags or the total number of instances affects accuracy, we performed experiments on how the number of labeled bags affects the accuracy in the case of overlap, i.e., when the total number of instances was fixed.
%\red{We prepared a fixed number of instances by randomly selecting 200 images per class from the dataset. }
%Then, we created bags with duplicates in various cases by changing the number of labeled bags (64, 128, 256, 512) with the bag size set to 10.
The initial bags were prepared, where the number of labeled bags was 64, and the bag size was 10. To increase the number of labeled bags, we then incrementally added new bags (64, 128, 256), where the new bags were generated from the initial instances that belong to the initial bags, i.e., generated bags overlap with the initial ones, as shown in Figure~\ref{fig:overlap} (c).
As shown in Figure\ref{fig:preliminary2}, the accuracy improves as the number of labeled bags increases in all datasets.
This observation is important for our {\em bag-level} data augmentation.
If we were to mimic this situation where the number of labeled bags increases and the total number of instances is fixed, it should improve accuracy, even though a situation with bags having overlaps is unnatural in actual applications. However, we can make this situation by generating new bags from the original ones.

\begin{figure*}[t]
    \centering
    \includegraphics[width=1.0\linewidth]{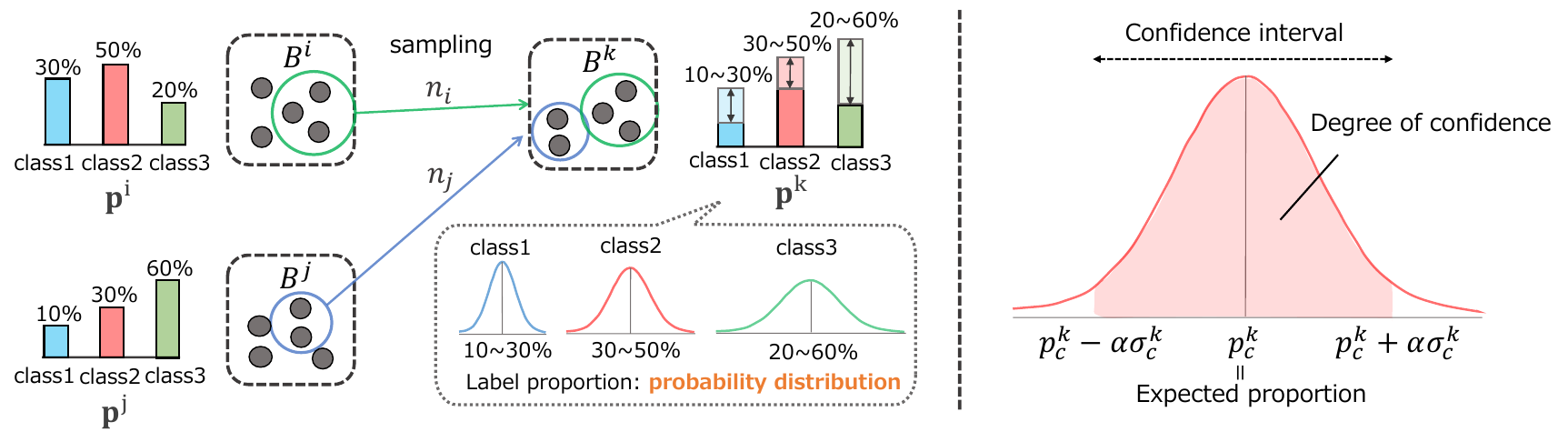}
    \caption{Overview of our method. {\bf Left}: the way to create mixed bags and illustration of label proportion's probability. {\bf Right}: confidence interval}
    \label{fig:mixbag}
\end{figure*}

\section{MixBag with confidence interval loss}
% MixBagのイメージ図 ＋ 信頼区間のイメージ図
To mimic the situation of increasing the number of labeled bags and allowing their overlap, we propose MixBag, which is a {\em bag-level} data augmentation technique.

\subsection{MixBag}
Figure \ref{fig:mixbag} (Left) shows an overview of MixBag.
Given two randomly selected bags $\{B^i, B^j\}$ where each bag is labeled with a proportion, $\bp_i$ or $\bp_j$, mix the two bags to generate a new bag.
More precisely, $n_i = |B^i| \times \gamma$ and $n_j = |B^j| \times (1-\gamma)$ instances are randomly sampled from $B^i$ and $B^j$, respectively, where $\gamma \in [0,1]$ is a random variable, and the sampled sub-bags are denoted as $S^i$ and $S^j$. Then, these two sub-bags are combined to generate a mixed bag, $B^{k}=S^i \cup S^j$.
This generated mixed bag overlaps with the original bags while the total number of instances is fixed, and the number of labeled bags is increased by the generated bags.
It can be considered to mimic the third preliminary experiment (Figure \ref{fig:overlap} (c)).

\subsection{Confidence interval loss}
The sampled sub-bags $S^i$, $S^j$ are expected to have almost the same proportions as the original ones, $\bp^i$ and $\bp^j$, if the randomly selected instances follow the proportion of the original bags. 
In this case, the expected proportion of the generated new bag $B^{k}$ is $\bp^{k} =\gamma \bp^i + (1-\gamma) \bp^j$.
However, in practice, the actual proportion of randomly sampled instances may have some gaps from that of the original bag because the number of sampled instances in each class does not always follow the original proportion.
This gap can be estimated by calculating the confidence interval (CI) for a population proportion.

Let us denote the proportion of a class $c$ in the $i$-th bag as $p_c^i$. The standard deviation of the class $c$ proportion of the sampled sub-bag $S^i$ is defined as:
% \begin{equation}
% \sigma_{p_c^i}^i = \sqrt{\frac{p_c^i(1-p_c^i)}{n_i}},
% \end{equation}
\begin{equation}
\sigma_{c}^i = \sqrt{\frac{p_c^i(1-p_c^i)}{n_i}},
\end{equation}
where $n_i = |B^i| \times \gamma$ is the number of instances randomly sampled from $B^i$. This equation is well-known in statistics; it can be derived using the central limit theorem.

The CI for the truth proportion $\tilde{p}_c^{i}$ of the sub-bag $S^i$ is formulated as:
% \begin{equation}
% p_c^i - \alpha \sigma_{p_c^i}^i \leq \hat{p}_c^i \leq p_c^i+ \alpha \sigma_{p_c^i}^i,
% \end{equation}
\begin{equation}
p_c^i - \alpha \sigma_{c}^i \leq \tilde{p}_c^i \leq p_c^i+ \alpha \sigma_{c}^i,
\end{equation}
where $\alpha$ is set according to the desired degree of confidence, e.g., 95\%.
If we set the confidence to a higher value, $\alpha$ becomes higher values.
For the other sub-bag $S_j$, the CI for the population proportion of $\tilde{p}_c^{j}$ is formulated in the same manner.

The CI for $\tilde{p}_c^{k}$, which is the proportion of the newly generated bag $B^{k}$, is defined as:
% \begin{align}
% r_c - \alpha \sigma_{r_c}^{k}\leq \hat{r_c} \leq r_c + \alpha \sigma_{r_c}^{k},\label{eq:CI}\\
% r_c = \gamma p_c^i + (1-\gamma) p_c^j, \label{eq:idealP}\\
\begin{align}
p_c^{k} - \alpha \sigma_{c}^{k}\leq \tilde{p}_c^{k} \leq p_c^{k} + \alpha \sigma_{c}^{k},\label{eq:CI}\\
p_c^{k} = \gamma p_c^i + (1-\gamma) p_c^j, \label{eq:idealP}\\
\gamma = \frac{n_i}{n_i + n_j}, \\
\sigma_{c}^{k} = \gamma \sqrt{\frac{p_c^i(1-p_c^i)}{n_i}} + (1-\gamma) \sqrt{\frac{p_c^j(1-p_c^j)}{n_j}},
%\sigma_{r_c}^{k} = \gamma \sigma_{p_c^i}^i + (1-\gamma) \sigma_{p_c^j}^j, \\
\label{eq:conf-int}
\end{align}
where $p_c^{k}$ is the expected proportion of the generated mixed bag, which is directly computed from the proportions of the original bags $B^i$, $B^j$, and the standard deviation $\sigma_{c}^{k}$ of the mixed bag can be calculated by using the standard deviation of the sub-bags with the ratio $\gamma$.

This confidence interval (Eq.\ref{eq:CI}) indicates that the actual proportion of the mixed bag has a gap from the expected proportion. If we directly use the proportion $\bp^{k}$ (Eq.\ref{eq:idealP}) for the mixed bag as augmented data and use it for the proportion loss, the gap from the actual proportion may adversely affect the training of the network since the gap causes a label noise.
%\red{the gap from the actual proportion may adversely affect the training of the network.}

To avoid this adverse effect caused by gaps (label noise), we propose a confidence interval loss (CI loss) for generated mixed bags.
The CI loss is based on the proportion loss~\cite{ardehaly2017co}.
In contrast to the standard proportion loss, the CI loss ignores a loss when the estimated proportion $\hat{p}_c^{k}$ is in the confidence interval $[p_c^{k} - \alpha \sigma_{c}^{k}, p_c^{k} + \alpha \sigma_{c}^{k}]$ since gaps in the CI often occur.

Given the degree of confidence $\alpha \%$ (the corresponding value is set to $\alpha$), such as when the degree of confidence is $95\%$, $\alpha=1.96$, the CI loss is defined as:
\begin{align}
    \ell_{\mathrm{CI}}(B^{k}, \bp^{k}, f) = - \sum_{c=1}^{C} \bm{1}(p_c^{k},\hat{p}_c^{k}) p_c^{k} \log \hat{p}_c^{k}, \\
    \hat{p}_c^{k} = \frac{1}{|B^{k}|} \sum_{j=1}^{|B^{k}|} f(x_j^{k})_c, \\
    % \bm{1}(p,p')=\{
    % \begin{array}{ll}
    % 0 & \text{if} p - \alpha \sigma\leq p' \leq p + \alpha \sigma\\
    % 1 & \text{otherwise}
    % \end{array}
    % \right,
\bm{1}(p,p')=
\left\{
\begin{array}{ll}
    0 & \text{if} \ p - \alpha \sigma\leq p' \leq p + \alpha \sigma\\
    1 & \text{otherwise}
\end{array}
\right.
\end{align}
where $\bm{1}$ is the indicator function, which outputs 0 if $p'$ is in the confidence interval, and 1 otherwise.
This CI loss updates the network parameters if the difference between the estimated proportion $\hat{p}_c^{k}$ and the expected proportion $p_c^{k}$ calculated by Eq.\ref{eq:idealP} is significant.
Using this loss, we can accelerate the training by using the augmented bags without any adverse effects from proportion gaps.

\begin{table*}[t]
    \centering
    \resizebox{\linewidth}{!}{%
        \begin{tabular}{|c|cccccccc||c|}
        \hline  
            {Method} & CIFAR10 & SVHN &PATH &OCT &BLOOD &ORGANA &ORGANC &ORGANS &Average\\    
            & Accuracy & Accuracy & Accuracy & Accuracy & Accuracy & Accuracy & Accuracy & Accuracy & Accuracy\\ \hline  \hline
            LLP~\cite{ardehaly2017co}   &0.4538  &0.3009 &0.7843 &0.4336 &0.8869 &0.7635 &0.7898 &0.5372&0.6187\\ 
            LLP + Ours(w/o CI)  &{0.4582} &{0.2971} &{0.7884}  &{0.425} &{0.8898}  &{0.7831} &{0.8189} &{0.563} &{0.6280}\\  
            \rowcolor{Gray}  LLP + Ours  &{0.5256} &{0.3742} &{0.7861}  &{0.4347} &{0.9017}  &{0.7971} &{0.8099} &{0.6197} &{\bf0.6561}\\ 
            LLP + Ours(supervised)  &{0.6594} &{0.7781} &{0.8118}  &{0.5396} &{0.8947}  &{0.8222} &{0.8233} &{0.6674} &{0.7234}\\ \hline 
            
            LLP-VAT~\cite{tsai2020}   &0.4740 &{0.3119} &{0.8915}  &{0.4124} &{0.8915}  &{0.7936} &{0.8060} &{0.5837} &{0.6455}\\  
            LLP-VAT + Ours(w/o CI)  &0.5203 &{0.4302} &{0.8779}  &{0.4503} &{0.8779}  &{0.7847} &{0.7656} &{0.5849} &{0.6614}\\  
            \rowcolor{Gray}  LLP-VAT + Ours &0.5283 &{0.4111} &{0.8944}  &{0.4366} &{0.8944}  &{0.8155} &{0.8228} &{0.6376} &{\bf 0.6800}\\  
            LLP-VAT + Ours(supervised) &0.5963 &{0.5976} &{0.8997}  &{0.4777} &{0.8997}  &{0.8230} &{0.8313} &{0.6622} &{0.7234}\\  \hline 
            LLP-PI~\cite{PI}   &0.4702  &0.3011 &0.8886   &0.4224  &0.8862 &{0.7810} &{0.7962} &{0.5917} &{0.6421}\\  
            LLP-PI + Ours(w/o CI)  &0.4794 &{0.5209} &{0.8688}  &{0.4350} &{0.8573}  &{0.7595} &{0.7570} &{0.5741} &{0.6565}\\  
            \rowcolor{Gray}  LLP-PI + Ours &0.5290 &{0.3865} &{0.8841}  &{0.4366} &{0.8841}  &{0.8013} &{0.7989} &{0.6236} &{\bf 0.6680}\\  
            LLP-PI + Ours(supervised) &0.5896 &{0.7721} &{0.8918}  &{0.4720} &{0.8918}  &{0.8101} &{0.8184} &{0.6577} &{0.7379}\\  \hline 
        \end{tabular}
    }
    \vspace{1mm}
    \caption{Instance-level classification accuracy when our method was applied to three baseline methods; LLP~\cite{ardehaly2017co}, LLP-VAT~\cite{tsai2020} and LLP-PI~\cite{PI}. Each value is the average of five-fold cross-validation results. Here, `CI' means the confident interval, and `supervised' indicates the case when a ground-truth proportion to the generated mixed bag is given.
    }
    \label{tab:eval-ours}
\end{table*}

\begin{table*}[t]
    \centering
    \resizebox{\linewidth}{!}{
    \begin{tabular}{|c|cccccccc||c|}
        \hline 
        {Method} & CIFAR10 & SVHN &PATH &OCT &BLOOD &ORGANA &ORGANC &ORGANS &Average\\    
        & Accuracy & Accuracy & Accuracy & Accuracy & Accuracy & Accuracy & Accuracy & Accuracy & Accuracy\\ \hline \hline 
        LLP~\cite{ardehaly2017co}   &0.4538  &0.3009 &0.7843 &0.4336 &0.8869 &0.7635 &0.7898 &0.5372 &0.6187\\
        \rowcolor{Gray} LLP + Ours  &0.5256 &0.3742 &0.7861  &0.4347 &0.9017  &0.7971 &0.8099 &{0.6197} &\bf{0.6561}\\ \hline 
        LLP + Flip   &0.5777  &0.3884  &0.8299  &0.4793  &0.9265  &0.6527 &0.6904 &0.6293 &0.6467\\
        \rowcolor{Gray} LLP + Flip + Ours  &0.5764 & 0.4124 &0.8005 &{0.5504} &0.9155 &0.6718 &0.6717 &0.6280 &{\bf0.6533}\\  \hline 
        LLP + Erase    &0.5551  &0.3500  &0.8097  &0.4414  &0.9094  &0.8393 &0.8447 &0.6578 &0.6759\\
        \rowcolor{Gray} LLP + Erase + Ours  &0.5671 &0.4269 &0.7934 &{0.4291} &0.9021 &0.8266 &0.8313 &0.6594 &{\bf0.6794}\\ \hline
        LLP + Invert   &0.5258  &0.6763  &0.7775  &0.4394  &0.8790 &0.7779 &0.8215 &0.6255 &0.6899\\
        \rowcolor{Gray} LLP + Invert + Ours  &0.5784 &0.7761 &0.7421 &{0.4580} &0.8736 & 0.7845 &0.8323 &0.6329 &{\bf0.7097}\\ \hline
        LLP + Gaussianblur  &0.4766  &0.3686  &0.7639  &0.4145 &0.8875  &0.7906  &0.8318 &0.6325 &0.6457\\
        \rowcolor{Gray} LLP + Gaussianblur + Ours  &0.5135 &0.4036 &0.7555 &0.4300 &0.8789 &0.7962 &0.7989 &0.6709 &{\bf0.6559}\\ \hline
        LLP + Persperctive  &0.5675  &0.5441  &0.7892  &0.4678 &0.8968  &0.8309  &0.8440  &0.6689  &0.7011\\
        \rowcolor{Gray} LLP + Perspective + Ours  &0.5750 &0.5715 &0.7745 &0.4962 &0.9064 &0.8180 &0.8404 &0.6895 &{\bf0.7089}\\ \hline
    \end{tabular}
    }
    \vspace{0.5mm}
    \caption{Instance-level classification accuracy when our method was applied after various instance-level augmentation methods.
    }
    \label{tab:comparison-aug}
\end{table*}

In training, mixed bags are randomly generated in each batch. It indicates that the number of generated bags will increase with iteration. We stop training based on the proportion loss in validation data. Note that instance-level labels cannot be used for validation in LLP.

\section{Experiments}
\subsection{Datasets and experimental settings}
\noindent
{\bf Preparation of bags:} 
The eight datasets used in these experiments were the same as those used in the preliminary experiments. The way of generating bags was also the same as in the preliminary experiments.
Although some of the related studies generated bags by allowing overlaps to increase the number of labeled bags, we did not allow overlaps since this situation is unnatural in actual applications.
To keep the experiment setups the same in all datasets, we set the number of labeled bags as 512 and the bag size as 10 based on the smallest dataset.
This small bag size can be considered a challenging scenario for MixBag since the number of samplings for sub-bags decreases, and it increases the confidence interval.
We also evaluated our method using different bag sizes to demonstrate the effectiveness of our method in various situations.

\vspace{5pt}

\noindent
{\bf Implementation:} We implemented our method by using Pytorch~\cite{paszke2019pytorch}. The network model was ResNet18~\cite{resnet} pre-trained on the ImageNet dataset~\cite{imagenet}.
To train our network, we also used the Adam Optimizer~\cite{resnet} with a learning rate of 3e-4, epoch$= 1000$, mini-batch size $= 32$, early stopping $= 10$ and trained on an NVIDIA GeForce 3090 GPU. We set the confidence degree as `99\%' and selected $\gamma$ randomly from a uniform distribution $[0,1]$.

\vspace{5pt}
\noindent
{\bf Evaluation metric:} We evaluated the model performance by the instance-level classification accuracies, which were calculated by using the confusion matrix.
To compute the metric, we performed a five-fold cross-validation in all our experiments. The accuracies listed in the tables are the average values of the five-fold cross-validation.

\subsection{Comparison}
\noindent
{\bf Comparison with current methods:} 
Our method MixBag can be applied to any proportion loss-based method since MixBag is a data augmentation method, in which any method can be applied after data augmentation with the CI loss.
Most current LLP methods incorporate self-supervised learning, such as consistency, into the proportion loss.
We thus applied our method in three methods: 1) LLP~\cite{ardehaly2017co}, which is the method of the original paper on using the proportion loss; 2) LLP-VAT~\cite{tsai2020}, which introduces the consistency regularization into the proportion loss; and 3) LLP-PI~\cite{PI}, which uses the additive Gaussian noise for making perturbed examples.
The LLP-VAT is one of the state-of-the-art LLP methods using the proportion loss, and its implementations are publicly available.

We compared the baseline methods as an ablation study. The first was `LLP', which uses the standard proportion loss; the second was `LLP+Ours(w/o CI),' which uses the MixBag without using the CI loss, where the proportion is directly calculated from the original bags (and may contain noise) was used as the label for the generated mixed bag. The third was `LLP+Ours(supervised)', which uses the ground-truth proportions of the mixed bags. Note that the proportion would be unknown in real cases.

Table \ref{tab:eval-ours} shows the accuracy of each method, where the gray color field indicates the proposed methods in each baseline method.
`LLP+Ours(w/o CI)' improved the accuracies in some datasets and slightly improved them on average from the baselines. However, the improvement was limited by the noisy proportions.
MixBag improved the accuracy of every baseline method on all datasets, and the improvement was 3 to 4 points on average.

\begin{table}[t]
    \centering
    \resizebox{0.75\linewidth}{!}{%
    \begin{tabular}{|c|c|c|}
    \hline  
        {Method}  &$\gamma$-sampling &Average Accuracy\\ \hline \hline 
         % & Accuracy\\ \hline  \hline
        LLP~\cite{ardehaly2017co}      &--        &0.6187\\ 
        LLP + Ours &uniform &{\bf0.6561}\\  
        LLP + Ours   &Gauss   &{0.6512}\\ 
        LLP + Ours    &half    &{0.6524}\\ \hline 
    \end{tabular}
    }
    \vspace{2mm}
    \caption{
    Average accuracy on eight datasets when changing a sampling method for $\gamma$.
    %'uniform' means uniform distribution. 'Gauss' means Gaussian distribution $\mathcal{N}(\mu=0.5,\sigma=0.25)$. `half' indicates that $\gamma$ is always 0.5.
    }
    \label{tab:generation}
\end{table}

\begin{table}[t]
    \centering
    \resizebox{0.7\linewidth}{!}{
        \begin{tabular}{|c|c|c|}
        \hline  
        {Method} &CI &Average Accuracy\\  \hline  \hline    
         % & Accuracy\\ \hline  \hline  
        LLP~\cite{ardehaly2017co}  & &0.6187\\ 
        LLP + Ours(99\%) &\checkmark   &{\bf0.6561}\\
        LLP + Ours(95\%) &\checkmark   &0.6519\\ 
        LLP + Ours(80\%) &\checkmark   &0.6531  \\ 
        LLP + Ours(50\%) &\checkmark   &0.6243 \\ \hline
        \end{tabular}
    }
    \vspace{2mm}
    \caption{Average accuracy on eight datasets when changing the degree of confidence interval (50\%, 80\%, 95\%, 99\%).
    %Accuracy in different confidence interval: . Each value is the average of 5fold cross-validation results. LLP + Ours means that we introduced MixBag into baseline LLP and we show our augmentation improve accuracy compared to LLP. Ours($n$\%) means $n$\% confidential interval of proposed method. Ours($99$\%) judge $1\%$ as noise which is out of confidential interval.
    }
    \label{tab:confidential-interval}
\end{table}

\vspace{5pt}
\noindent
{\bf Introducing MixBag with instance-level augmentation}:
Next, we introduced our method to the instance-level augmentation methods.
We used standard augmentation techniques: `Flip' randomly flips the image horizontally and upside down. `Erase' removes a randomly selected area from an original image. `Invert' randomly changes the colors of the image by inversion. `Gaussianblur' blurs an image by adding a randomly chosen Gaussian blur. `Perspective' performs a random projective transformation on an image.
To generate a bag by instance-level augmentation, we added perturbations into instances randomly selected from an original bag; i.e., an augmented bag contains the perturbated instances and original instances, where this augmentation has not been reported in previous papers.
In LLP + instance-level + MixBag, the mixed bag was created from the original and instance-level augmented bags.
%A new bag was generated from the augmented instances in which the bag had the same proportion as the original one.
%As in the above comparison, MixBag can also be applied to any instance-level data augmentation; i.e., it can be applied based on the augmented data.
Table \ref{tab:comparison-aug} shows the accuracies of the baseline and our methods. The instance-level data augmentation improved the performance from the baseline. Our method further improved the accuracies of every instance-level augmentation method.

\subsection{Performance in various situations}
%We next evaluate how our method works robustly in various cases.

\noindent
{\bf Performance with different $\gamma$ sampling methods}:
We evaluated our method with three different $\gamma$ sampling methods, i.e., three different ways to determine the mixing rate $\gamma$ of the two sub-bags $S^i, S^j$: `uniform' selects $\gamma$ randomly from a uniform distribution; 
`Gauss' selects $\gamma$ randomly from a Gaussian distribution $\mathcal{N}(0.5, 0.25)$; `half' always set $\gamma$ to 0.5.
Table \ref{tab:generation} shows the average accuracies of all datasets. 
All sampling methods outperformed the performance compared to the baseline method (LLP); `uniform' was marginally better than other methods.

\begin{table}[t]
    \centering
    \resizebox{0.7\linewidth}{!}{%
        \begin{tabular}{|c|cc||c|}
        \hline  
             {Method} &Size &Num &Average\\   
            &&& Accuracy\\ \hline  \hline

            LLP~\cite{ardehaly2017co}   &10 &512 &0.6187\\ 
            LLP + Ours(w/o CI)  &10 &512 &{0.6280}\\  
            \rowcolor{Gray}  LLP + Ours &10 &512 &{\bf0.6561}\\ 
            \hline
            %LLP + Ours(supervised) &10 &512  &{0.6594} &{0.7781} &{0.8118}  &{0.5396} &{0.8947}  &{0.8222} &{0.8233} &{0.6674} &{0.7234}\\ \hline  

            % \multicolumn{10}{|c|}{Bag size = 20, The number of bag = 256}\\ \hline
            LLP~\cite{ardehaly2017co} &20 &256 &0.5441\\ 
            LLP + Ours(w/o CI) &20 &256  &0.5496\\  
            \rowcolor{Gray}  LLP + Ours &20 &256  &{\bf 0.5667}\\ 
            \hline
            %LLP + Ours(supervised) &20 &256  &0.6330  &0.7272 &0.7967 &0.4582 &0.8893 &0.7954 &0.8048 &0.6263&0.7164\\ \hline  

            % \multicolumn{10}{|c|}{Bag size = 40, The number of bag = 128}\\ \hline
            LLP~\cite{ardehaly2017co} &40 &128 &0.4746 \\ 
            LLP + Ours(w/o CI) &40 &128 &0.4600\\  
            \rowcolor{Gray}  LLP + Ours  &40 &128 &{\bf 0.5058}\\ 
            \hline 
        \end{tabular}
    }
    \vspace{2mm}
    \caption{
    Average accuracy on eight datasets when changing the number of labeled bags and bag size. 'Size' means the bag size. 'Num' means the number of labeled bags.
    }
    \label{tab:ablation-differentdata}
\end{table}

\vspace{5pt}
\noindent
{\bf Performance with different confidence intervals}: 
We conducted experiments with four different confidence intervals, $99\%$, $95\%$, $80\%$, and $50\%$.
Table \ref{tab:confidential-interval} shows the average accuracies of all datasets.
For all confidence interval settings,  our method improved the accuracy on all datasets and in all cases. When the confidence interval took high values ($99\%$, $95\%$, $80\%$), our method worked well since the worst effects of noisy labels were mitigated by the CI loss.
Even when the confidence interval was low ($50\%$), the accuracy improved from the baseline method.
The case of $99\%$ was the best on average.
We consider that a larger confidence interval avoids the adverse effects of noisy proportions and makes the training more stable.

\vspace{5pt}
\noindent
{\bf Performance when changing the number of labeled bags and bag size}:
To show the effectiveness of our method in various situations, we conducted experiments with different bag sizes (512, 256, 128) and different numbers of labeled 
bags (10, 20, 40), where the bag size and the number of labeled bags have a trade-off since the total number of instances is fixed.
Table \ref{tab:ablation-differentdata} shows the average accuracies on all datasets with various setups. Our method improved the accuracy from the baseline methods on average in all cases.
This demonstrates the robustness of our method.

\vspace{5pt}
\noindent
{\bf Performance with different bag generations}:
To confirm the effectiveness of the `Mix' approach, we compared MixBag with two other bag generation methods.
1) 'union' directly combines any two bags ($B^i, B^j$) without sampling. In this case, the proportion of a generated bag can be directly calculated from those of the original bags; i.e., there is no noise for the combined proportions by taking the union of two bags. Therefore, the CI loss was not used for this method.
2) `sub-bag' randomly samples instances from a single bag $B^i$. It can be considered as a sub-bag of the original bag $B^i$.
In this case, the label proportions of the sub-bag may have a gap from the original one, and thus, we evaluated two cases; `sub-bag' without and with the CI loss.
3) `MixBag' is our method, which mixes two bags. We also evaluated two cases; Mixbag with and without CI loss.

Table \ref{tab:ablation-bag-generation} shows the average of the instance-level accuracies of the comparative methods on the eight datasets.
%`sub-bag w/o CI' decreased the accuracy because of the sampling noise for the label proportion in a sub-bag.
Since `sub-bag with CI' and union' can increase the number of labeled bags, these methods had slightly improved accuracy from that of baseline method LLP~\cite{ardehaly2017co}.
Our method, `MixBag with CI' further improved the accuracy compared with these other bag generation methods.
We consider that our method has two characteristics of `sub-bag' and `union', where MixBag first makes sub-bags from original bags and then takes the union of these sub-bags. In addition, MixBag can generate various proportions of the mixed bag compared to `sub-bag' and `union'. Therefore, MixBag was better than them.

\begin{table}[t]
    \centering
    \resizebox{0.8\linewidth}{!}{%
        \begin{tabular}{|c|c|c|c|}
        \hline
             \multirow{2}{*}{Method} &\multirow{2}{*}{Bag-Generation} &\multirow{2}{*}{CI} &Average\\ 
             & & &Accuracy\\ 
             \hline  \hline
            LLP~\cite{ardehaly2017co} &-- &-- &0.6187 \\ 
            LLP + Ours(w/o CI)  &Union  &--  &0.6226 \\
            LLP + Ours(w/o CI)  &Sub-bag & &0.4956 \\
            LLP + Ours(with CI) &Sub-bag &\checkmark &0.6364 \\
            LLP  + Ours(w/o CI)  &MixBag   & &0.6280 \\ 
            \rowcolor{Gray}LLP + Ours(with CI)   &MixBag &\checkmark &{\bf0.6561}\\ 
            \hline
        \end{tabular}}
    \vspace{2mm}
    \caption{Average accuracy on eight datasets in different bag-generation methods.
    %Ablation study: Accuracies of in different bag-generation methods. 'Union' directly combines two bags $B^i$ and $B^j$ without sampling. 'Sub-bag' picks out instances from the bag $B^i$ and generates a sub-bag $S^i$. 'Mix' means mixing two sub-bags $S^i$ and $S^j$ together.
    }
    \label{tab:ablation-bag-generation}
\end{table}
\subsection{Detailed Analysis}
\begin{figure}[t]
    \begin{tabular}{c}
    \centering
        \begin{minipage}{0.48\linewidth}
            \centering
            \includegraphics[width=1.0\linewidth]{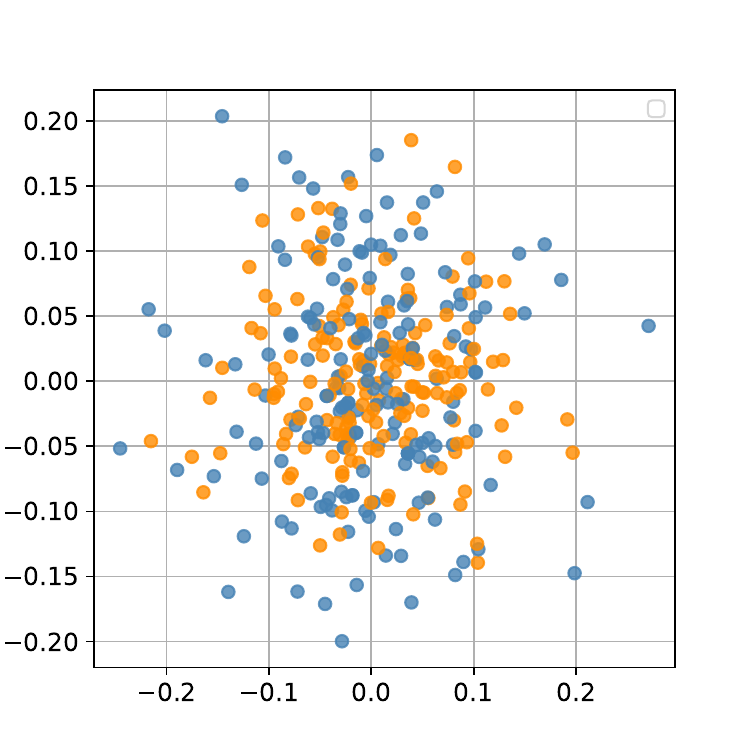}
            \caption{Distribution of proportion vectors. {\bf Blue}: Original bag's proportion. {\bf Orange}: Mixed bag's proportion.
            %Proportion distribution in proportion label space. {\bf Blue}: Original bag's proportion. {\bf Orange}: MixBag's proportion.
            }
            \label{fig:scatter}
        \end{minipage}%
        \begin{minipage}{0.02\textwidth}
            \hfill
        \end{minipage}
        \begin{minipage}{0.48\linewidth}
            \centering
            \includegraphics[width=1.0\linewidth]{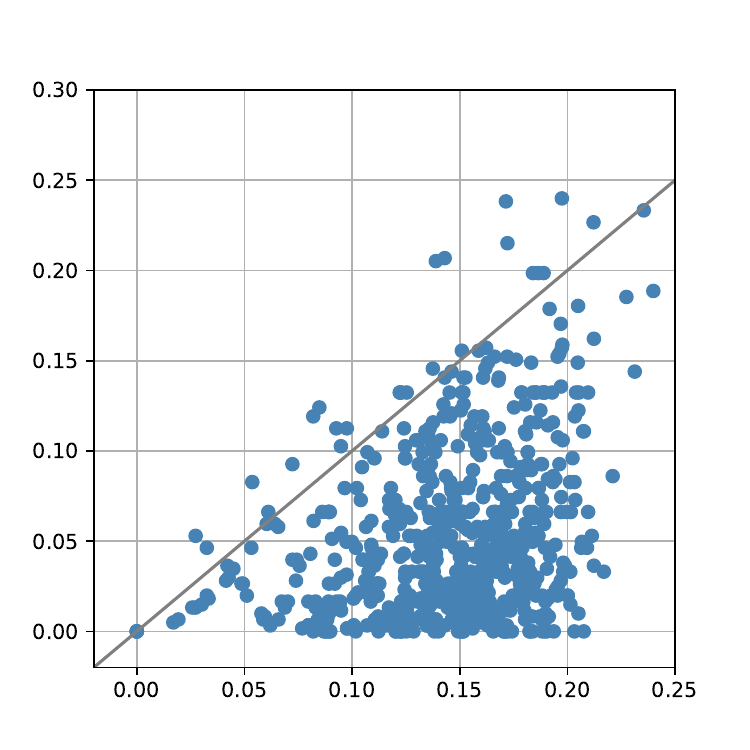}
            \caption{Relationship between confidence interval and actual gaps. {\bf Blue}: proportion vector of a bag.
            %Relationship between confidential interval and ground-truth propotion. {\bf Blue}: Mixbag's proportion of each class.
            }
            \label{fig:analysis}
        \end{minipage}
    \end{tabular}
\end{figure}

%〇詳細な評価
%・Proportionラベル空間でのProportionベクトルの分布
%　オリジナルと、Augmentation後
%・正解のProportionと信頼区間の関係性
%　正解Proportion と 推定Proportion（平均）の差 を横軸、信頼区間の幅を縦軸で散布図を記載
%・特徴量分布の推移
%　学習が進むにつれて、どう特徴空間が変わっていくか
%（この辺は松尾君の実験とも共通するところがありそう）
\noindent
%\red{{\bf Distribution of proportion vectors in proportion label space}: }
{\bf Distribution of proportion vectors in the space of proportion labels}:
Figure \ref{fig:scatter} shows distributions of the original proportion vectors $\{\bp^i\}_{i=1}^n$ (blue) and the generated mixed bag's proportion vectors $\{\br^i\}_{i=1}^m$ (orange) in the `ORGANS' dataset, in which the dimension was compressed to 2D by PCA~\cite{wold1987principal}.
Here, the number of mixed bags is the same as that of the original bags. 
We can see that the MixBag proportion vectors do not overlap with the original ones, i.e., new proportion labels were generated, and the distribution of mixed bag proportions covers the original distribution by interpolating pairs of the original bags.
This result shows that MixBag adds diversity to the training dataset, which facilitates an improvement in the instance-level classification accuracy.

\vspace{5pt}
\noindent
{\bf Relationship between confidence interval and ground-truth proportion}: 
Figure \ref{fig:analysis} shows how well the confidence interval works.
The horizontal axis indicates the difference ($\|\bp - \hat{\bp}\|_1$) between the ground-truth proportion $\bp$ and the estimated proportion $\hat{\bp}$ of a mixed bag.
The vertical axis indicates the difference ($\alpha\bm{\sigma}^{k}$) between the upper bounds of the confidence interval and $\bp$, where we set the degree of confidence as 99\%.
Each blue point indicates the proportion vector of a mixed bag.
Almost all points are under the straight line in the figure. This indicates that the actual proportion gaps from the ground truth were less than the confidence interval, and thus the proposed CI loss works properly.

% given sets of unlabeled observations, each set with known label
% proportions, predict the labels of another set of observations, possibly with known label proportions

\section{Conclusion}
In this paper, we examined how the number of labeled bags and the bag size affects the performance in LLP. As a result, we found that the accuracy improves as the number of labeled bags increases, even when the total number of instances is fixed. Then, based on this observation, we propose a simple but effective {\em bag-level} data augmentation method. In addition, we also proposed a confidence interval loss that effectively trains a classification network by using the generated bags while avoiding the adverse effects caused by noisy proportions.
To the best of our knowledge, this is the first attempt to propose {\em bag-level} data augmentation for LLP.
The experiments using eight datasets demonstrated the effectiveness of our method in various cases. Additionally, MixBag can be applied to instance-level data augmentation techniques and any LLP method that uses a standard proportion loss.

\vspace{5pt}
\noindent
{\bf Acknowledgements}: This work was supported by JSPS KAKENHI Grant Number JP20H04211 and JP23K18509, and JST, ACT-X Grant Number JPMJAX200G, Japan. \\

{\small
\bibliographystyle{ieee_fullname}
\bibliography{refer}

\begin{thebibliography}{10}\itemsep=-1pt

\bibitem{ardehaly2017co}
Ehsan~Mohammady Ardehaly and Aron Culotta.
\newblock Co-training for demographic classification using deep learning from
  label proportions.
\newblock In {\em 2017 IEEE International Conference on Data Mining Workshops
  (ICDMW)}, pages 1017--1024, 2017.

\bibitem{chen2020gridmask}
Pengguang Chen, Shu Liu, Hengshuang Zhao, and Jiaya Jia.
\newblock Gridmask data augmentation.
\newblock {\em arXiv preprint arXiv:2001.04086}, 2020.

\bibitem{cubuk2019autoaugment}
Ekin~D Cubuk, Barret Zoph, Dandelion Mane, Vijay Vasudevan, and Quoc~V Le.
\newblock Autoaugment: Learning augmentation strategies from data.
\newblock In {\em Proceedings of the IEEE/CVF conference on computer vision and
  pattern recognition}, pages 113--123, 2019.

\bibitem{cubuk2020randaugment}
Ekin~D Cubuk, Barret Zoph, Jonathon Shlens, and Quoc~V Le.
\newblock Randaugment: Practical automated data augmentation with a reduced
  search space.
\newblock In {\em Proceedings of the IEEE/CVF conference on computer vision and
  pattern recognition workshops}, pages 702--703, 2020.

\bibitem{imagenet}
Jia Deng, Wei Dong, Richard Socher, Li-Jia Li, Kai Li, and Li Fei-Fei.
\newblock Imagenet: A large-scale hierarchical image database.
\newblock In {\em 2009 IEEE conference on computer vision and pattern
  recognition}, pages 248--255. Ieee, 2009.

\bibitem{mnist}
Li Deng.
\newblock The mnist database of handwritten digit images for machine learning
  research [best of the web].
\newblock {\em IEEE signal processing magazine}, 29(6):141--142, 2012.

\bibitem{devries2017improved}
Terrance DeVries and Graham~W Taylor.
\newblock Improved regularization of convolutional neural networks with cutout.
\newblock {\em arXiv preprint arXiv:1708.04552}, 2017.

\bibitem{DulacArnoldG2020}
Dulac-Arnold Gabriel, Zeghidour Neil, Cuturi Marco, Beyer Lucas, and Vert
  Jean-Philippe.
\newblock Deep multi-class learning from label proportions.
\newblock {\em arXiv preprint arXiv:1905.12909}, 2020.

\bibitem{resnet}
Kaiming He, Xiangyu Zhang, Shaoqing Ren, and Jian Sun.
\newblock Deep residual learning for image recognition.
\newblock In {\em Proceedings of the IEEE conference on computer vision and
  pattern recognition}, pages 770--778, 2016.

\bibitem{hendrycks2019augmix}
Dan Hendrycks, Norman Mu, Ekin~D Cubuk, Barret Zoph, Justin Gilmer, and Balaji
  Lakshminarayanan.
\newblock Augmix: A simple data processing method to improve robustness and
  uncertainty.
\newblock {\em arXiv preprint arXiv:1912.02781}, 2019.

\bibitem{inoue2018data}
Hiroshi Inoue.
\newblock Data augmentation by pairing samples for images classification.
\newblock {\em arXiv preprint arXiv:1801.02929}, 2018.

\bibitem{cifar10}
Alex Krizhevsky, Geoffrey Hinton, et~al.
\newblock Learning multiple layers of features from tiny images.
\newblock 2009.

\bibitem{PI}
Samuli Laine and Timo Aila.
\newblock Temporal ensembling for semi-supervised learning.
\newblock {\em arXiv preprint arXiv:1610.02242}, 2016.

\bibitem{li2020fencemask}
Pu Li, Xiangyang Li, and Xiang Long.
\newblock Fencemask: a data augmentation approach for pre-extracted image
  features.
\newblock {\em arXiv preprint arXiv:2006.07877}, 2020.

\bibitem{li2021novel}
Zekun Li, Wei Zhao, Feng Shi, Lei Qi, Xingzhi Xie, Ying Wei, Zhongxiang Ding,
  Yang Gao, Shangjie Wu, Jun Liu, et~al.
\newblock A novel multiple instance learning framework for covid-19 severity
  assessment via data augmentation and self-supervised learning.
\newblock {\em Medical Image Analysis}, 69:101978, 2021.

\bibitem{lim2019fast}
Sungbin Lim, Ildoo Kim, Taesup Kim, Chiheon Kim, and Sungwoong Kim.
\newblock Fast autoaugment.
\newblock {\em Advances in Neural Information Processing Systems}, 32, 2019.

\bibitem{liu2019learning}
Jiabin Liu, Bo Wang, Zhiquan Qi, Yingjie Tian, and Yong Shi.
\newblock Learning from label proportions with generative adversarial networks.
\newblock {\em Advances in neural information processing systems}, 32, 2019.

\bibitem{ijcai2021p377}
Jiabin Liu, Bo Wang, Xin Shen, Zhiquan Qi, and Yingjie Tian.
\newblock Two-stage training for learning from label proportions.
\newblock In {\em Proceedings of the Thirtieth International Joint Conference
  on Artificial Intelligence, {IJCAI-21}}, pages 2737--2743, 8 2021.

\bibitem{matsuoICASSP2023}
Shinnosuke Matsuo, Ryoma Bise, Seiichi Uchida, and Daiki Suehiro.
\newblock Learning from label proportion with online pseudo-label decision by
  regret minimization.
\newblock In {\em Proceedings of the IEEE International Conference on
  Acoustics, Speech and Signal Processing (ICASSP)}, 2023.

\bibitem{svhn}
Yuval Netzer, Tao Wang, Adam Coates, Alessandro Bissacco, Bo Wu, and Andrew~Y
  Ng.
\newblock Reading digits in natural images with unsupervised feature learning.
\newblock 2011.

\bibitem{paszke2019pytorch}
Adam Paszke, Sam Gross, Francisco Massa, Adam Lerer, James Bradbury, Gregory
  Chanan, Trevor Killeen, Zeming Lin, Natalia Gimelshein, Luca Antiga, et~al.
\newblock Pytorch: An imperative style, high-performance deep learning library.
\newblock In {\em NeurIPS}, pages 8026--8037, 2019.

\bibitem{perez2017effectiveness}
Luis Perez and Jason Wang.
\newblock The effectiveness of data augmentation in image classification using
  deep learning.
\newblock {\em arXiv preprint arXiv:1712.04621}, 2017.

\bibitem{qi2016learning}
Zhiquan Qi, Bo Wang, Fan Meng, and Lingfeng Niu.
\newblock Learning with label proportions via npsvm.
\newblock {\em IEEE transactions on cybernetics}, 47(10):3293--3305, 2016.

\bibitem{rueping2010svm}
Stefan Rueping.
\newblock Svm classifier estimation from group probabilities.
\newblock In {\em Proceedings of the 27th international conference on machine
  learning (ICML-10)}, pages 911--918, 2010.

\bibitem{shorten2019survey}
Connor Shorten and Taghi~M Khoshgoftaar.
\newblock A survey on image data augmentation for deep learning.
\newblock {\em Journal of big data}, 6(1):1--48, 2019.

\bibitem{tang2020unified}
Hao Tang, Hong Liu, and Nicu Sebe.
\newblock Unified generative adversarial networks for controllable
  image-to-image translation.
\newblock {\em IEEE Transactions on Image Processing}, 29:8916--8929, 2020.

\bibitem{tokunagaECCV2020}
Hiroki Tokunaga, Brian~Kenji Iwana, Yuki Teramoto, Akihiko Yoshizawa, and Ryoma
  Bise.
\newblock Negative pseudo labeling using class proportion for semantic
  segmentation in pathology.
\newblock In {\em Computer Vision--ECCV 2020: 16th European Conference,
  Glasgow, UK, August 23--28, 2020, Proceedings, Part XV 16}, pages 430--446.
  Springer, 2020.

\bibitem{tsai2020}
Kuen-Han Tsai and Hsuan-Tien Lin.
\newblock Learning from label proportions with consistency regularization.
\newblock In {\em Asian Conference on Machine Learning}, pages 513--528. PMLR,
  2020.

\bibitem{wold1987principal}
Svante Wold, Kim Esbensen, and Paul Geladi.
\newblock Principal component analysis.
\newblock {\em Chemometrics and intelligent laboratory systems}, 2(1-3):37--52,
  1987.

\bibitem{yang2021two}
Haoran Yang, Wanjing Zhang, and Wai Lam.
\newblock A two-stage training framework with feature-label matching mechanism
  for learning from label proportions.
\newblock In {\em Asian Conference on Machine Learning}, pages 1461--1476,
  2021.

\bibitem{yang2022remix}
Jiawei Yang, Hanbo Chen, Yu Zhao, Fan Yang, Yao Zhang, Lei He, and Jianhua Yao.
\newblock Remix: A general and efficient framework for multiple instance
  learning based whole slide image classification.
\newblock In {\em Medical Image Computing and Computer Assisted
  Intervention--MICCAI 2022: 25th International Conference, Singapore,
  September 18--22, 2022, Proceedings, Part II}, pages 35--45. Springer, 2022.

\bibitem{medmnistv2}
Jiancheng Yang, Rui Shi, Donglai Wei, Zequan Liu, Lin Zhao, Bilian Ke,
  Hanspeter Pfister, and Bingbing Ni.
\newblock Medmnist v2-a large-scale lightweight benchmark for 2d and 3d
  biomedical image classification.
\newblock {\em Scientific Data}, 10(1):41, 2023.

\bibitem{ShiY2020}
Shi Yong, Liu Jiabin, Wang Bo, Qi Zhiquan, and Tian YingJie.
\newblock Deep learning from label proportions with labeled samples.
\newblock pages 73--81, 2020.

\bibitem{yun2019cutmix}
Sangdoo Yun, Dongyoon Han, Seong~Joon Oh, Sanghyuk Chun, Junsuk Choe, and
  Youngjoon Yoo.
\newblock Cutmix: Regularization strategy to train strong classifiers with
  localizable features.
\newblock In {\em Proceedings of the IEEE/CVF international conference on
  computer vision}, pages 6023--6032, 2019.

\bibitem{zhang2017mixup}
Hongyi Zhang, Moustapha Cisse, Yann~N Dauphin, and David Lopez-Paz.
\newblock mixup: Beyond empirical risk minimization.
\newblock {\em arXiv preprint arXiv:1710.09412}, 2017.

\bibitem{zhu2017unpaired}
Jun-Yan Zhu, Taesung Park, Phillip Isola, and Alexei~A Efros.
\newblock Unpaired image-to-image translation using cycle-consistent
  adversarial networks.
\newblock In {\em Proceedings of the IEEE international conference on computer
  vision}, pages 2223--2232, 2017.

\end{thebibliography}
}

%%%%%%%%%%%%%%%%%%%%%%%%%%%%%%%%%%%%%%%%%%%%%%

\begin{table*}[t]
\centering
\resizebox{\linewidth}{!}{%
\begin{tabular}{|c|c|cccccccc||c|}
\hline  
    {Method} &$\gamma$-sampling & CIFAR10 & SVHN &PATH &OCT &BLOOD &ORGANA &ORGANC &ORGANS &Average\\    
    & &Accuracy & Accuracy & Accuracy & Accuracy & Accuracy & Accuracy & Accuracy & Accuracy & Accuracy\\ \hline  \hline
    LLP  &-- &0.4538  &0.3009  &0.7843 &0.4336 &0.8869 &0.7635 &0.7898 &0.5372 &0.6187\\ 
    LLP + Ours &uniform  &{0.5256} &{0.3742} &{0.7861}  &{0.4347} &{0.9017}  &{0.7971} &{0.8099} &{0.6197} &{\bf0.6561}\\  
    LLP + Ours &gauss  &{0.5162} &{0.3710} &{0.7853}  &{0.4307} &{0.8980}  &{0.7867} &{0.7937} &{0.6283} &{0.6512}\\ 
    LLP + Ours &half  &{0.5060} &{0.3774} &{0.7819}  &{0.4400} &{0.8912}  &{0.8050} &{0.8015} &{0.6164} &{0.6524}\\ \hline \end{tabular}}
\vspace{0.5mm}
\caption{Average accuracy on eight datasets when changing a sampling method for $\gamma$. Note that this table corresponds to Table 3 in the main paper, which shows the average accuracy of all datasets. All sampling methods outperformed the performance compared to the baseline method (LLP); `uniform' was marginally better than other methods.}
\label{tab:generation}
\end{table*}

\begin{table*}[t!]
    \centering
    \resizebox{\linewidth}{!}{
        \begin{tabular}{|c|cccccccc||c|}
        \hline  
        {Method} & CIFAR10 & SVHN &PATH &OCT &BLOOD &ORGANA &ORGANC &ORGANS &Average\\   
        & Accuracy & Accuracy & Accuracy & Accuracy & Accuracy & Accuracy & Accuracy & Accuracy & Accuracy\\ \hline  \hline  
        LLP  &0.4538  &0.3009  &0.7843 &0.4336 &0.8869 &0.7635 &0.7898 &0.5372 &0.6187\\ 
        LLP + Ours(99\%)  &{0.5256} &{0.3742} &{0.7861}  &{0.4347} &{0.9017}  &{0.7971} &{0.8099} &{0.6197} &{\bf0.6561}\\
        LLP + Ours(95\%)  &{0.5267} &{0.3807} &{0.7913}  &0.4227 &{0.8852}  &{0.8034} &{0.8012} &{0.6046} &0.6519\\ 
        LLP + Ours(80\%)  &{0.5069} &{0.4015} &{0.8004}  &0.4004 &0.8947  &{0.8104} &0.7970 &0.6142 &0.6531  \\ 
        LLP + Ours(50\%)  &{0.4781} &{0.3493} &{0.7611}  &0.4019 &{0.8676}  &{0.7794} &0.8008 &{0.5562} &0.6243 \\ \hline
        \end{tabular}
    }
    \vspace{0.5mm}
    \caption{
    Accuracy on each dataset when changing the degree of confidence interval (50\%, 80\%, 95\%, 99\%).
    Note that this table corresponds to Table 4 in the main paper, which shows the average accuracy of all datasets.
    LLP + Ours(99\%) was superior to the baseline LLP in all datasets.
    % Accuracy performance in different confidential interval: . Each value is the average of 5fold cross-validation results. LLP + Ours means that we introduced mixbag into baseline LLP and we show our augmentation improve accuracy compared to LLP. Ours($n$\%) means $n$\% confidential interval of proposed method. Ours($99$\%) judge $1\%$ as noise which is out of confidential interval.
    }
    \label{tab:confidential-interval}
\end{table*}

\begin{table*}[ht]
    \centering
    \resizebox{\linewidth}{!}{%
        \begin{tabular}{|c|cc|cccccccc||c|}
        \hline  
             {Method} &Size &Num &CIFAR10 &SVHN &PATH &OCT &BLOOD &ORGANA &ORGANC &ORGANS &Average\\   
            &&& Accuracy & Accuracy & Accuracy & Accuracy & Accuracy & Accuracy & Accuracy & Accuracy & Accuracy\\ \hline  \hline

            LLP   &10 &512 &0.4538  &0.3009  &0.7843 &0.4336 &0.8869 &0.7635 &0.7898 &0.5372 &0.6187\\ 
            LLP + Ours(w/o CI)  &10 &512 &{0.4582} &{0.2971} &{0.7884}  &{0.4250} &{0.8898}  &{0.7831} &{0.8189} &{0.563} &{0.6280}\\  
            \rowcolor{Gray}  LLP + Ours &10 &512  &{0.5256} &{0.3742} &{0.7861}  &{0.4347} &{0.9017}  &{0.7971} &{0.8099} &{0.6197} &{\bf0.6561}\\ 
            \hline
            %LLP + Ours(supervised) &10 &512  &{0.6594} &{0.7781} &{0.8118}  &{0.5396} &{0.8947}  &{0.8222} &{0.8233} &{0.6674} &{0.7234}\\ \hline  

            % \multicolumn{10}{|c|}{Bag size = 20, The number of bag = 256}\\ \hline
            LLP &20 &256 &0.3301 &0.2217 &0.7339 &0.3778 &0.8527  &0.6936 &0.7034 &0.4396 &0.5441\\ 
            LLP + Ours(w/o CI) &20 &256  &0.3286  &0.2217  &0.7223 &0.4080 &0.8650 &0.7064 &0.7270 &0.4182  &0.5496\\  
            \rowcolor{Gray}  LLP + Ours &20 &256  &0.3705  &0.2305 &0.7521 &0.3876 &0.8687 &0.7312 &0.7435 &0.4499  &{\bf 0.5667}\\ 
            \hline
            %LLP + Ours(supervised) &20 &256  &0.6330  &0.7272  &0.7967 &0.4582 &0.8893 &0.7954 &0.8048 &0.6263  &0.7164\\ \hline  

            % \multicolumn{10}{|c|}{Bag size = 40, The number of bag = 128}\\ \hline
            LLP &40 &128 &0.2727 &0.2042  &0.6870 &0.3790 &0.8061 &0.5669 &0.5422 &0.3386 &0.4746  \\ 
            LLP + Ours(w/o CI) &40 &128 &0.2579  &0.2141 &0.6968 &0.351 &0.7995 &0.5357 &0.5059 & 0.3191 &0.4600\\  
            \rowcolor{Gray}  LLP + Ours  &40 &128 &0.3162  &0.2034  &0.7002 &0.3836 &0.8167 &0.6249 &0.6168 &0.3847  &{\bf 0.5058}\\ 
            %LLP + Ours(supervised)  &40 &128 &0.5272  &0.5940 &0.7705 &0.3992 &0.8706 &0.7502 &0.7812 &0.5787  &0.6590\\ 
            % \hline 

            % LLP \cite{LLP} &80 &64 & &  &  &  & & & & &  \\ 
            % LLP + Ours(w/o CI) &80 &64 & &  &  &  & & & & &\\  
            % \rowcolor{Gray}  LLP + Ours  &80 &64 & &  &  &  & & & & &\\ 
            %LLP + Ours(supervised)  &40 &128 &0.5272  &0.5940 &0.7705 &0.3992 &0.8706 &0.7502 &0.7812 &0.5787  &0.6590\\ 
            \hline 
        \end{tabular}
    }
    \vspace{0.5mm}
    \caption{Average accuracy on eight datasets when changing the number of labeled bags and bag size. 'Size' means the bag size. 'Num' means the number of labeled bags. Note that this table corresponds to Table 5 in the main paper, which shows the average accuracy of all datasets. LLP + Ours was superior to the baseline LLP in all datasets under every bag size and the number of labeled bags.}
    \label{tab:ablation-differentdata}
\end{table*}

\begin{table*}[t]
    \centering
    \resizebox{\linewidth}{!}{%
        \begin{tabular}{|c|c|c|cccccccc||c|}
        \hline
             \multirow{2}{*}{Method} &\multirow{2}{*}{Bag-Generation} &\multirow{2}{*}{CI} &CIFAR10 &SVHN &PATH &OCT &BLOOD &ORGANA &ORGANC &ORGANS &Average\\ 
             & & &Accuracy&Accuracy&Accuracy&Accuracy&Accuracy&Accuracy&Accuracy&Accuracy&Accuracy\\ 
             \hline  \hline
            LLP &-- &-- &0.4538  &0.3009  &0.7843 &0.4336 &0.8869 &0.7635 &0.7898 &0.5372 &0.6187 \\ 
            LLP + Ours(w/o CI)  &Union  &--  &0.4575 &0.3252 &0.7710 &0.4014 &0.8882 &0.7884 &0.7887 &0.5600 &0.6226 \\
            LLP + Ours(w/o CI)  &Sub-bag & &0.3155 &0.2331 &0.6221 &0.4182 &0.7587 &0.6035 &0.5836 &0.4305 &0.4956 \\
            LLP + Ours(with CI) &Sub-bag &\checkmark &0.5157 &0.3785 &0.7853 &0.4294 &0.8882 &0.7906 &0.8109 &0.6033 &0.6364 \\
            LLP  + Ours(w/o CI)  &MixBag   & &{0.4582} &{0.2971} &{0.7884}  &{0.4250} &{0.8898}  &{0.7831} &{0.8189} &{0.5630} &{0.6280} \\ 
            \rowcolor{Gray}LLP + Ours(with CI)   &MixBag &\checkmark &{0.5256} &{0.3742} &{0.7861}  &{0.4347} &{0.9017}  &{0.7971} &{0.8099} &{0.6197} &{\bf0.6561}\\ 
            \hline
        \end{tabular}}
    \vspace{2mm}
    \caption{Accuracy on each dataset in different bag generation methods. Note that this table corresponds to Table 6 in the main paper, which shows the average accuracy of all datasets. LLP + Ours(MixBag with CI) was superior to the baseline LLP in almost all datasets under every condition.
    %Ablation study: Accuracies of in different bag-generation methods. 'Union' directly combines two bags $B^i$ and $B^j$ without sampling. 'Sub-bag' picks out instances from the bag $B^i$ and generates a sub-bag $S^i$. 'Mix' means mixing two sub-bags $S^i$ and $S^j$ together.
    }
    \label{tab:ablation-bag-generation}
\end{table*}

\end{document}